# 자기 주의 메커니즘을 위한 다중 트랜스포머를 사용한 이미지 캡셔닝

파루크 올리모브*°, 시카 두베*, 러비나 스레스터, 뜨란 뜨릉 띤, 전문구

전저전기컴퓨터공학

광주과학기술원

olimov.farrukh@gist.ac.kr, shikha.d@gm.gist.ac.kr, labina.shr@gm.gist.ac.kr, ttrungtin@gm.gist.ac.kr, mgjeon@gist.ac.kr

# Image Captioning using Multiple Transformers for Self-Attention Mechanism

Farrukh Olimov*°, Shikha Dubey*, Labina Shrestha, Tran Trung Tin, Moongu Jeon

School of Electrical Engineering and Computer Science

Gwangju Institute of Science and Technology

## Abstract (요약)

Real-time image captioning, along with adequate precision, is the main challenge of this research field. The present work, Multiple Transformers for Self-Attention Mechanism (MTSM), utilizes multiple transformers to address these problems. The proposed algorithm, MTSM, acquires region proposals using a transformer detector (DETR). Consequently, MTSM achieves the self-attention mechanism by transferring these region proposals and their visual and geometrical features through another transformer and learns the objects' local and global interconnections. The qualitative and quantitative results of the proposed algorithm, MTSM, are shown on the MSCOCO dataset.

## 1. Introduction

Neural Image Captioning is a challenging task that combines cutting-edge approaches from computer vision and natural language processing. With the prevailing popularity of RNN models, many models based on RNNs were introduced [1, 2, 3, 5, 6, 8, 9, 14] and achieved a significant performance boost in the image captioning task. Recently, encoder-decoder architecture is in great demand; therefore, the algorithm proposed by [1] plays a fundamental role in fascinating people to the image captioning research field. Afterward, more complex architectures were proposed, such as the method proposed in [10], which used complex architecture like Faster-RCNN for region proposals and utilized these regions for attention mechanism.

Moreover, real-time image captioning, along with adequate precision, is the main challenge of this research field. Therefore, our present work utilizes multiple transformers to address these problems due to its recent advancement in machine translation [12] and computer vision tasks [11]. The proposed algorithm, MTSM, acquires region proposals using a transformer detector (DETR) [11] and achieves the self-attention mechanism [10] by transferring these region proposals and their visual and geometrical features through another transformer and learns the objects' local and global interconnections.

## 2. The Proposed Image Captioning Algorithm

### 2.1. Object Detection

* Equal Contribution

In the proposed algorithm (Figure 1.), MTSM, we have utilized the newly introduced transformer-based object detection method, DETR [11], for the object region proposals as compared to the recently proposed method of Image Captioning that uses Faster-RCNN for object proposal network. The main difference between Faster-RCNN and Transformer Network are as follow:

a) Faster-RCNN trains RPN and RoI layers separately by taking anchors as inputs for RPN. The output propagates down to the RoI layer, resulting in many possible bounding boxes produced per object. To have a single bounding box per image, Non-Maximum-Suppression is used, which discards overlapping bounding boxes that exceed some fixed threshold. All

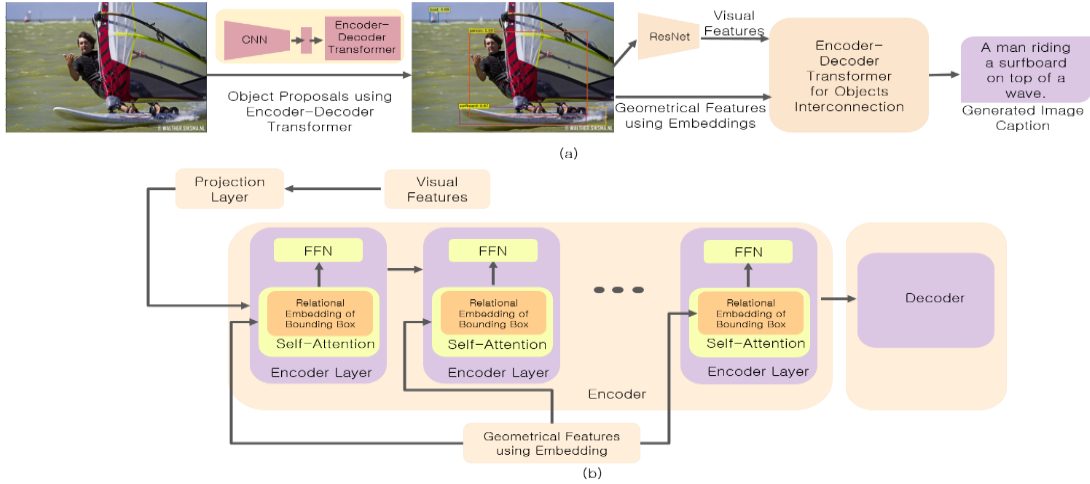

Figure 1. (a) Architecture of the proposed Image Captioning Method, MTSM. (b) A detailed structure of the Objects Interconnection Transformer.

the above steps can distort output due to human intervention in any step.

b) On the other hand, DETR [11] has end-to-end training, and no suppression mechanism is used for prediction. It is based on standard encoder-decoder architecture, with a slight modification of decoder, predicting N objects' bounding box coordinates and corresponding probability. According to the original paper, DETR's performance at least similar and, in some cases, overtakes Faster-RCNN. Moreover, DETR is faster in computation than Faster-RCNN.

Therefore, we have utilized a transformer detector for region proposals. In the proposed algorithm, MTSM, we have extracted bounding boxes with the probability greater or equal to 0.7. Later, N extracted objects, and their bounding boxes were passed to ResNet-50 to get the visual feature maps of the 2048 dimension and then projected down to 512-dimensional vector using trainable matrix W before propagating through encoder layers (Figure 1.a).

2.2 Transformer Model

We followed the transformer model introduced by Vaswani et al. [12], which has L identical layers composed of 2 sub-layers: multi-head self-attention and position-wise, fully connected layers.

Decoder and Encoder, as shown in Figure 1.b, with geometrical self-attention specified by [10] in given in Equation 1.

Geometric attention is calculated as follows:

$$\theta_G^{np} = ReLU(Emb(\delta)W_G) \quad (1)$$

and then incorporated into original attention weights as follows:

$$\theta^{np} = \frac{\theta_G^{np} exp(\theta_A^{np})}{\sum N_{l=1} \theta_G^{nl} exp(\theta_A^{nl})} \quad (2)$$

where $\delta$ represents $log_{10}$ (center coordinates, width, height). So, each self-attention will be modified according to following Equation (3) as follows:

$$head(Q,K,V,\theta) = softmax\left(\theta_g exp\left(\frac{QK^T}{\sqrt{(dk)}}\right)\right)V \quad (3)$$

3. Implementation and Evaluation

3.1. Dataset

We trained our model on the MSCOCO [13] dataset with around 113k images for the training set and each 5k images for validation and test set. We used minor preprocessing of captions, namely lowercased and removed punctuations. To keep consistency with other models and for the lightweight property, we limit the vocabulary to the words that occur at least 5 times in the whole corpus, which produces a vocabulary size of 10,041 words. However, the maximum length of a sentence is fixed to at most 51 words.

### 3.2. Implementation Details

The model is build using Tensorflow v2 on a machine with Nvidia 1080 Ti and RAM 16GB. The model was trained for 30 epochs with masked XE objective using Adam optimizer. Unlike other transformer-based models [10, 11, 12], the step-wise learning rate is changed to epoch-wise learning decays with the initial learning rate at $1e^{-5}$. The maximum number of objects is fixed to 78 objects. Moreover, the number of layers in each Encoder and decoder of the transformer is L=6.

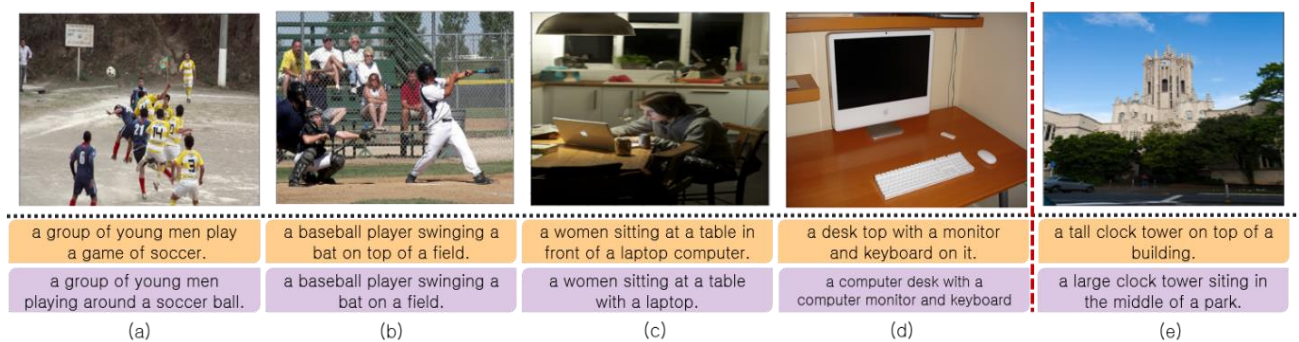

Figure 2. Qualitative Analysis of the proposed algorithm on the MSCOCO 2014 testing dataset, where 🟧 color boxes and 🟪 color boxes represent ground truth captions and predicted captions, respectively. (a) – (b) represent successful cases of image caption prediction, and (e) represents the case where our method fails to predict meaningful caption because of the massive occlusion of the object.

### 3.3. Evaluation

We have evaluated the proposed algorithm, MTSM, based on BLEU metrics and illustrated BLEU score 1-4 in Table 2. The proposed algorithm, MTSM, is evaluated and compared with previously proposed algorithms is presented in Table 1. using BLEU-4 score and our model outperforms mostly mentioned algorithms. Moreover, we have provided qualitative analysis in Figure 2, which shows the performance of MTSM on images from the test set.

### 4. Conclusion and Future Work

This paper has proposed an algorithm, MTSM, which exploited transformers in image captioning tasks and outperforms other algorithms on the MSCOCO test dataset. As modern architectures are trained in a reinforcement manner, our future work will include further training of the model using BLEU-4 or other metrics as a reward function similar to [10]. They trained and reported their highest BLEU to score using a reinforcement manner.

#### Acknowledgement

This work was supported by Institute of Information Communications Technology Planning Evaluation (IITP) grant funded by the Korea Government (MSIT) (No. 2014-3-00077, Development of Global Multitarget Tracking and Event Prediction Techniques Based on Real-time Large-scale Video Analysis).

Table 1. BLEU Score Comparison on MSCOCO 2014 Testing Dataset. MIXER-B stands for MIXER without a Baseline [8] (Karpathy Splits).

| Algorithms | BLEU-4 |
|---|---|
| Hard-Attention [2] | 24.3 |
| Embedding Reward [7] | 30.4 |
| SCST: Att2all [9] | 30.0 |
| SCST: Att2in [9] | 31.3 |
| MIXER-B [8] | 30.9 |
| MIXER [8] | 31.7 |
| Google NIC [1] | 32.1 |
| PG-SPIDEr [5] | 32.2 |
| LSTM-A [4] | 32.5 |
| Adaptive [3] | 33.2 |
| Actor-Critic [6] | 34.4 |
| ORT (w/o R.L.) [10] | 35.5 |
| MTSM (Ours) | 35.9 |

Table 2. Performance Analysis of the Proposed Method, MTSM using BLEU Scores 1-4 on MSCOCO 2014 Testing Dataset.

| Algorithm | BLEU-1 | BLEU-2 | BLEU-3 | BLEU-4 |
|---|---|---|---|---|
| MTSM (Ours) | 75.3 | 58.9 | 46.4 | 35.9 |

References:

[1] Oriol Vinyals *et al.*, "Show and tell: A neural image caption generator," CVPR, 2015.

[2] Kelvin Xu *et al.*, "Show, attend and tell: Neural image caption generation with visual attention," ICML, 2015.

[3] Jiasen Lu *et al.*, "Knowing when to look: Adaptive attention via a visual sentinel for image captioning," CVPR, 2017.

[4] Ting Yao *et al.*, "Boosting image captioning with attributes," OpenReview, 2016.

[5] Siqi Liu *et al.*, "Improved image captioning via policy gradient optimization of spider," arXiv preprint arXiv:1612.00370, 2016.

[6] Li Zhang *et al.*, "Actor-Critic Sequence Training for Image Captioning," arXiv preprint arXiv:1706.09601, 2017.

[7] Zhou Ren *et al.*, "Deep reinforcement learning-based image captioning with embedding reward," arXiv preprint arXiv:1704.03899, 2017.

[8] Marc'Aurelio Ranzato *et al.*, "Sequence level training with recurrent neural networks," ICLR, 2015.

[9] S. J. Rennie *et al.*, "Self-critical sequence training for image captioning," CVPR, 2017.

[10] Herdade Simao *et al.*, "Image Captioning: Transforming Objects into Words," NeurIPS, 2019.

[11] Carion Nicolas *et al.*, "End-to-End Object Detection with Transformers," ECCV, 2020.

[12] Vaswani Ashish *et al.*, "Attention Is All You Need," NIPS, 2017.

[13] T.-Y. Lin *et al.*, "Microsoft COCO: Common objects in context," ECCV, 2014.

[14] A. Karpathy and L. Fei-Fei, "Deep visual-semantic alignments for generating image descriptions," CVPR, 2015.